\documentclass[journal,twoside,web]{ieeecolor}
\usepackage{tmi}
\usepackage{cite}
\usepackage{amsmath,amssymb,amsfonts}
\usepackage{algorithmic}
\usepackage{graphicx}
\usepackage{textcomp}
\usepackage{fontawesome}
\usepackage{algorithm}
\usepackage{mdframed}
\usepackage{caption}
\usepackage{tabularx}

\usepackage[table, xcdraw, dvipsnames]{xcolor}
\colorlet{red}{black}
\colorlet{blue}{black}
\newcommand{\revB}[1]{\textcolor{blue}{#1}}
\definecolor{keywordcolor}{RGB}{178,34,34}
\definecolor{cadetblue}{RGB}{95,158,160}
\def\BibTeX{{\rm B\kern-.05em{\sc i\kern-.025em b}\kern-.08em
    T\kern-.1667em\lower.7ex\hbox{E}\kern-.125emX}}
\usepackage{array,booktabs,multirow,makecell}
\usepackage{colortbl}
\usepackage{arydshln}
\usepackage{bm}

\newcolumntype{I}{!{\vrule width 1pt}}
\makeatletter
\newcommand{\thickhline}{%
    \noalign {\ifnum 0=`}\fi \hrule height 1pt
    \futurelet \reserved@a \@xhline
}
\makeatother
\definecolor{mygray}{gray}{.9}
\usepackage[pagebackref=true,breaklinks=true,colorlinks,bookmarks=true]{hyperref}

\newcommand{\selft}{{Multimodal Methods}}

\newcommand{\repft}{{Image Only}}

\definecolor{mydarkdarkred}{RGB}{182, 58, 43}
\definecolor{mydarkdarkred2}{RGB}{180, 92, 67}
\definecolor{mydarkred}{RGB}{245, 220, 215}
\definecolor{mylightred}{RGB}{251, 244, 242}

\definecolor{mydarkdarkblue}{RGB}{17, 24, 129}
\definecolor{mydarkdarkblue2}{RGB}{92, 108, 165}
\definecolor{mydarkblue}{RGB}{211, 218, 234}
\definecolor{mylightblue}{RGB}{241, 244, 250}

\definecolor{mydarkdarkgreen}{RGB}{93, 150, 74}
\definecolor{mydarkgreen}{RGB}{216, 233, 199}
\definecolor{mylightgreen}{RGB}{245, 249, 241}

\definecolor{darkred}{RGB}{139,0,0}

\definecolor{mydarkyellow}{RGB}{216, 214, 196}
\definecolor{mylightyellow}{RGB}{245, 245, 240} 

\definecolor{mygray}{gray}{.9}
\definecolor{mygreen}{RGB}{93,173,85}
\definecolor{myorange}{RGB}{233,144,61}
\definecolor{azure}{rgb}{0.0, 0.5, 1.0}
\definecolor{gray}{rgb}{0.3, 0.3, 0.3}
\definecolor{DarkGreen}{RGB}{42,110,63}
\definecolor{DarkBlue}{RGB}{64,101,149}

\definecolor{mylightgray}{RGB}{249, 249, 249}
\usepackage{hyperref}
\usepackage{cleveref}
\usepackage{pifont}

\definecolor{mygray}{gray}{.9}
\definecolor{mygreen}{RGB}{93,173,85}
\definecolor{myorange}{RGB}{233,144,61}
\definecolor{azure}{rgb}{0.0, 0.5, 1.0}
\definecolor{gray}{rgb}{0.3, 0.3, 0.3}
\definecolor{DarkGreen}{RGB}{42,110,63}
\definecolor{DarkBlue}{RGB}{64,101,149}
\definecolor{mylightgray}{RGB}{249, 249, 249}

\newcommand{\tworowup}[2]{%
  #1\,{\tiny\textcolor{myorange}{\ensuremath{\pm}\,\textbf{#2}}}%
}
\newcommand{\tworowdown}[2]{%
  #1\,{\tiny\textcolor{mygreen}{\ensuremath{\pm}\,\textbf{#2}}}%
}

\begin{document}
\title{Anatomy-aware Fine-grained Multimodal Fusion for Laryngopharyngeal Cancer T-Staging Prediction Using CT and Radiology Report}
\author{Xingyue Zhao, Yanzhou Su, Fang Zhang, Zhanghexuan Ji, Yirui Wang, Dazhou Guo, Sibo Ju, Yuehua Cheng, Yuzhen Chen, Ming Feng, Le Lu, \IEEEmembership{Fellow, IEEE}, Tsung-Ying Ho, Jian Wang,  Dakai Jin, \\ \IEEEmembership{Member, IEEE}, Na Shen 
\thanks{X. Zhao, Y. Su and F. Zhang contributed equally to this work. Corresponding authors: J. Wang (wang.jian2@zs-hospital.sh.cn), D. Jin (Dakai.jin@gmail.com) and N. Shen (shen.na@zs-hospital.sh.cn).}
\thanks{X. Zhao, Y. Su, Z. Ji, Y. Wang, D. Guo, L. Lu and D. Jin are with DAMO Academy, Alibaba Group. X. Zhao and M. Feng are with Department of Neurosurgery, Peking Union Medical College Hospital, Chinese Academy of Medical Sciences and Peking Union Medical College, Beijing, China. Y. Su is also with Hupan Laboratory, Hangzhou, China. S. Ju is with Fuzhou University, Fuzhou, China. F. Zhang is with Department of Radiology, Eye \& ENT Hospital, Fudan University, Shanghai, China. J. Wang is with Department of Radiology, Zhongshan Hospital, Fudan University, Shanghai, China. Y. Chen and T-Y. Ho is with Department of Nuclear Medicine, Chang Gung Memorial Hospital, Taiwan, ROC. Y. Cheng and N. Shen are with Department of Otolaryngology-Head \& Neck Surgery, Zhongshan Hospital, Fudan University, Shanghai, China. N. Shen is also with Department of Otolaryngology, Zhongshan Hospital(Xiamen), Fudan University, Xiamen, China.}
\thanks{This work was supported by the Shanghai Natural Science Foundation of China (25ZR1401055), the AI for Science Foundation of Fudan University (X24AI059), the Fujian Provincial Department of Science and Technology Guidance Project (2025D033) and Xuhui District Academy-Local Cooperation Project of Shanghai(25XHYD-09).}
}

\maketitle

\begin{abstract}
Accurate T-staging is crucial for guiding personalized treatment strategies for laryngopharyngeal cancer. However, current clinical practice relies on invasive biopsy procedures, whereas CT-based staging remains challenging due to the complex patterns of tumor invasion. Recent computer-aided approaches face two key challenges: 1) Structural relationship modeling: existing methods underrepresent anatomically structured patterns of tumor invasion, as they either process whole CT volumes without tumor-specific anatomical constraints or rely on labor-intensive tumor segmentation. 2) Fine-grained cross-modal alignment: while radiology reports contain organ-specific invasion details, current methods that apply global feature fusion struggle to accurately align individual anatomical structures with their corresponding textual descriptions. To address these issues, we propose an anatomy-aware multimodal framework that integrates organ-level CT context and radiology reports into a unified representation for laryngopharyngeal T-staging. The framework first constructs an \textcolor{red}{Anatomy-Structured Organ Graph (AOG)} that captures invasion patterns between primary sites and surrounding organs, then performs \textcolor{red}{Organ-Anchored Cross-Modal Alignment (OCA)} so that each organ node aggregates textual evidence from the radiology report, and finally refines this graph representation by injecting organ-specific invasion cues extracted from the report via \textcolor{red}{Report-Enhanced Graph-Refinement (REG)}, yielding a multimodal organ graph that combines spatial and textual evidence. Extensive experiments demonstrate that the proposed framework achieves superior performance in T-staging of laryngopharyngeal cancer.
\end{abstract}

\begin{IEEEkeywords}
Tumor staging, multimodal fusion, CT, radiology report, laryngopharyngeal cancer.
\end{IEEEkeywords}

\section{Introduction}
\label{sec:introduction}
\begin{figure}[!t]
\centerline{\includegraphics[width=\columnwidth]{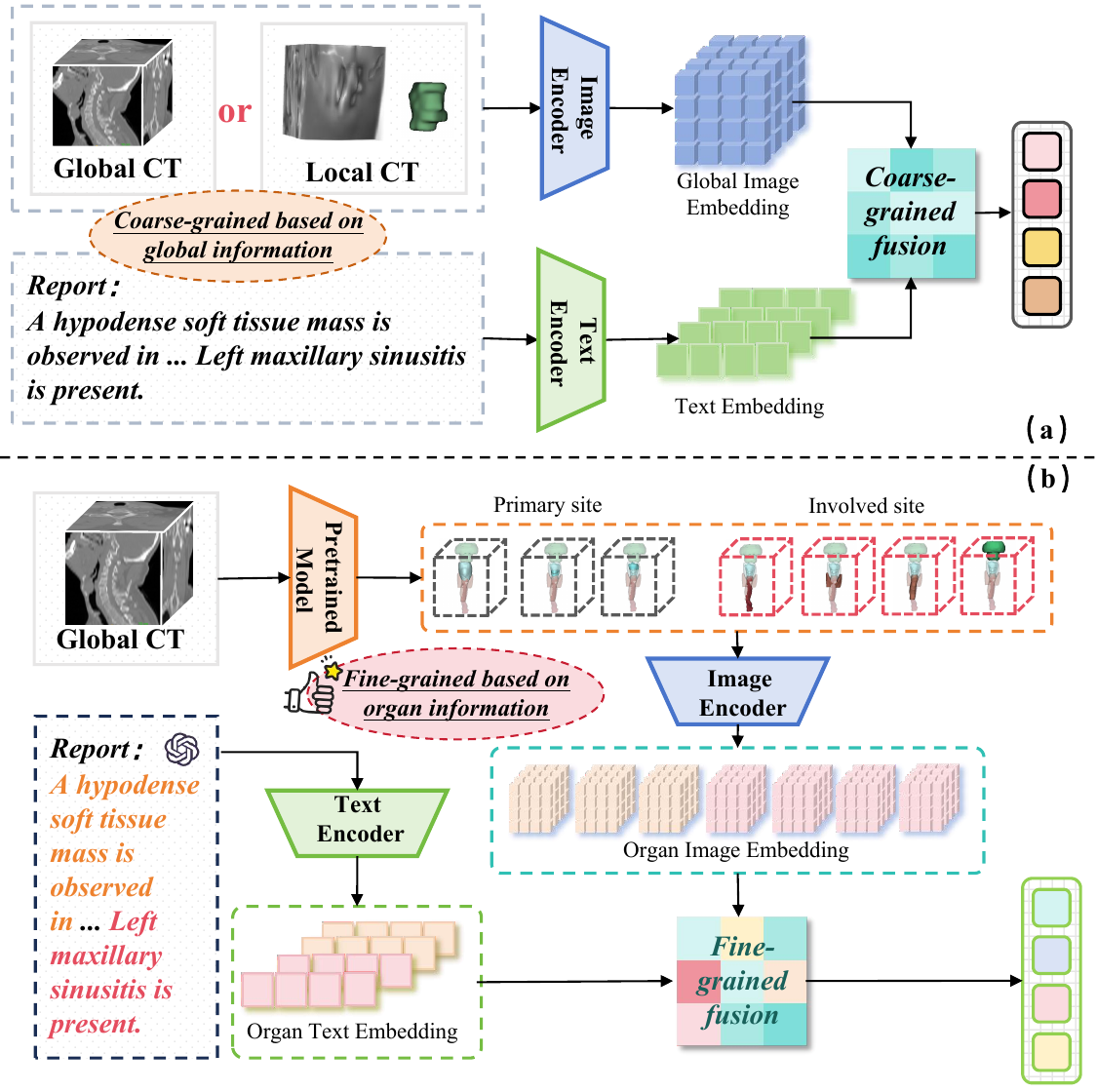}}
\caption{Comparison between (a) conventional coarse-grained image–text fusion based on global CT and report information \revB{\cite{jain2024mmbcd,guo2024common,bui2024visual,li2023lvit,lee2023text}}, and (b) the proposed anatomy-structured multimodal framework.}%
\label{fig1}
\end{figure}

Accurate T-staging in laryngopharyngeal cancer is pivotal for clinical decision-making, as treatment strategies differ markedly between T stages~\cite{chen2023combined,guo2020ct,ren2018magnetic}. Early-stage disease (T1–T2) is usually managed with partial laryngectomy, laser resection, or definitive radiotherapy, whereas advanced stages (T3–T4) often require total laryngectomy combined with adjuvant therapies~\cite{kuno2014primary,shen2026pretreatment}. Although the current reference standard relies on biopsy-derived pathology, this process is invasive and time-consuming. Direct CT-based T-staging is therefore attractive as a non-invasive and rapid alternative, but manually reviewing an entire 3-D CT volume remains labor-intensive and subjective: radiologists must track the tumor across tens of axial slices and judge its extension into adjacent organs, and staging accuracy can suffer, especially in borderline cases.

Recent progress in computer-aided diagnosis has encouraged researchers to predict T-stage directly from CT scans. Radiomics-based models have been developed to discriminate between adjacent T-stages (e.g., T2/T3 or T3/T4)~\cite{liu2024machine,wang2019radiomic}, and convolutional neural networks (CNNs) have been used to assess thyroid cartilage invasion, an essential indicator of advanced-stage disease~\cite{santin2019detecting,hao2024deep}. However, these approaches remain unimodal, relying solely on imaging features and ignoring the clinical information documented in routine radiology reports. Given the success of multimodal image–text learning in other medical applications and the absence of such methods for laryngopharyngeal cancer, there is a need for T-staging models that jointly exploit CT scans and their paired radiology reports.

Unlike conventional image classification tasks, T-staging in laryngopharyngeal cancer presents unique challenges that demand anatomy-aware modeling. The larynx is a particularly complex region in head and neck imaging, with numerous critical structures tightly packed into a small space. Tumors typically arise from a primary site and extend into neighboring organs, so staging depends more on this invasion pattern than the tumor in isolation. Nevertheless, existing computational approaches struggle in two key aspects (see Figure \ref{fig1}). First, \textbf{structural relationship modeling}: many methods either process whole CT volumes without tumor annotations or rely on labor-intensive tumor segmentation, making it difficult to explicitly represent inter-organ invasion patterns. Second, \textbf{fine-grained cross-modal alignment}: although radiology reports describe clinical findings, current multimodal models usually fuse image and text at a global level (e.g., via cross-attention or feature concatenation)~\revB{\cite{jain2024mmbcd,guo2024common,bui2024visual,li2023lvit,lee2023text}}, hindering precise association between specific anatomical structures and their relevant clinical descriptions. These gaps motivate a fundamental question: \textbf{\textit{how can we integrate organ-level CT context and radiology reports into a unified, anatomy-structured multimodal representation for laryngopharyngeal T-staging?}}

To capture the characteristic pattern of tumor spread in laryngopharyngeal cancer, we first construct an \textbf{Anatomy-Structured Organ Graph (AOG)} that explicitly models spatial relationships among key laryngopharyngeal structures. Rather than operating on whole CT volumes or relying on manual tumor delineation, AOG leverages a publicly available pre-trained organ segmentation model~\cite{ji2023continual} to obtain reliable organ-specific features, taking advantage of the relatively consistent appearance of anatomical structures. Guided by clinical prior knowledge, it distinguishes primary tumor sites from potentially invaded organs and instantiates an anatomy-guided connectivity pattern: primary sites are fully interconnected to encode their internal relationships and are further linked to organs that are likely to be invaded. The resulting organ-level graph representation describes tumor spread as structured interactions between primary sites and surrounding organs, providing an anatomy-aware basis for downstream T-staging.
 
\textit{Once organ-level structural relationships are captured by our anatomy-structured organ graph (AOG), incorporating textual evidence via fine-grained image–text alignment becomes crucial for comprehensive staging assessment.} To this end, we introduce an \emph{Organ-Anchored Cross-Modal Alignment (OCA)} module. OCA provides cross-modal interaction at the organ level: each organ node derived from CT serves as a query over the radiology report and selectively aggregates fine-grained textual evidence that is most relevant to its appearance, location, and suspected involvement, yielding text-enhanced organ representations while keeping the anatomy-guided graph topology unchanged. This design naturally raises a question: since radiology reports often contain explicit statements about which organs are invaded, why not construct the organ graph directly from these documented invasion relationships instead of linking all anatomically at-risk organs? 

In practice, however, radiologists typically report only obvious and clinically salient invasion, hence subtle or equivocal extension may be omitted. Relying solely on reported invasion would therefore disconnect organs that are anatomically at risk but not explicitly mentioned, leaving their CT features underutilized. To address this, we retain the anatomy-driven connectivity in AOG and use report information as a high-confidence but incomplete guidance signal. Specifically, we design a \emph{Report-Enhanced Graph-Refinement  (REG)} module that reuses the global report representation as a text node in the graph, employs a large language model (LLM) to identify which organs are explicitly described as invaded in the report, and connects this text node only to primary sites and to those invaded organs, while keeping the original organ–organ edges unchanged. This produces a report-guided organ–text graph in which anatomy-based risk relationships are maintained but selectively strengthened along pathways supported by the report. The resulting multimodal organ graph is then fed into a graph neural network that jointly exploits anatomical and report-derived information to predict laryngopharyngeal T-stage.

\noindent\textbf{Our contributions are summarized as follows:}
\begin{itemize}
\item[{\ding{182}}] \textbf{An anatomy-guided, annotation-free multimodal T-staging framework.}
We propose, to our knowledge, the first multimodal laryngopharyngeal cancer T-staging framework that constructs an organ-level graph from pre-trained organ segmentations and integrates paired radiology reports without requiring manual tumor masks, making it practical for full 3D volumes in routine clinical workflows.

\item[{\ding{183}}] \textbf{Reframing T-staging as anatomy-structured multimodal prediction.}
We cast laryngopharyngeal cancer T-staging as prediction over an anatomy-structured organ graph that combines CT-derived organ features with organ-level textual cues. By focusing on inter-organ relationships and invasion patterns rather than treating the tumor in isolation, our model is designed to capture key patterns of disease involvement across critical laryngopharyngeal structures.

\item[{\ding{184}}] \textbf{Comprehensive empirical validation on a multi-center cohort.}
We conduct extensive experiments on a cohort of 656 patients from four hospitals (to our knowledge, the largest laryngopharyngeal T-staging dataset with paired CT and radiology reports), in both binary (T1–T2 vs.\ T3–T4) and three-class settings, and evaluate generalization on internal and external test sets, demonstrating consistent improvements over strong imaging-only and multimodal baselines.
\end{itemize}

\section{Related Work}
\subsection{T-staging}
T-staging is a key component of cancer staging, describing the size, extent, and invasion depth of the primary tumor. Accurate staging is vital for diagnosis, treatment planning, and prognosis evaluation. CT, MRI, PET and endoscopic ultrasound are widely applied in clinical practice. Recent advances in deep learning have spurred interest in automating T-staging of cancers based on various imaging modalities.
For rectal cancer, Han et al.~\cite{han2023convolutional} employ a convolutional neural network (CNN) to predict T-stage from CT, capturing subtle infiltration cues. Zhang et al.~\cite{zhang2024colorectal} propose a tensor graph network exploiting spatial--structural information for improved segmentation-based staging. Dong et al.~\cite{dong2024cross} develop a cross-domain mutual-assistance framework for nasopharyngeal cancer, and further introduced interpretable dual-branch networks (IPNet) for endoscopic ultrasound-based rectal T-staging.
In prostate cancer, Ogino et al.~\cite{ogino2021augmented} present a feature space transfer model using MRI to address domain adaptation and enhance robustness. For head and neck cancers, Gilbert et al.~\cite{gilbert2010staging} compare standard and dedicated laryngopharyngeal CT for staging accuracy, while Kuno et al.~\cite{kuno2014primary} review CT, MRI, and dual-energy CT applications. Liu et al.~\cite{liu2024machine} build a radiomics-based classifier to distinguish T2 from T3 in laryngopharyngeal cancer (LHSCC), and Chen et al.~\cite{chen2023combined} achieve improved laryngopharyngeal cancer staging via multimodal fusion of radiomics and deep learning using CT. \revB{These laryngopharyngeal and head-and-neck staging studies, however, are single-center, limited in scale (118--319 patients), and rely on imaging features alone without exploiting paired radiology reports.}
Despite the progress, laryngopharyngeal cancer staging remains challenging due to complex invasion patterns, limited CT resolution and contrast, and insufficient structural modeling in current algorithms. We address these gaps with a multimodal framework that integrates anatomical relationship modeling and textual evidence, aiming to enhance fine-grained invasion characterization and T-staging accuracy.

\subsection{Multi-modal Learning}
Multi-modal learning leverages complementary information from diverse data modalities and has shown strong potential in clinical diagnosis and medical image analysis~\cite{lian2025efficient,chen2025holodx}. By integrating medical images with textual reports, such models capture richer semantic representations, improving both diagnostic accuracy and interpretability~\cite{liu2023improving,hamamci2026generalist}. Recent studies have explored vision–language integration for medical imaging. TP-DRSeg~\cite{li2024tp} injects medical prior knowledge from text into a vision-only segmentation network, while TextDiff~\cite{feng2024enhancing} enriches diffusion models with text annotations to align semantic features and descriptions. SimTxtSeg~\cite{xie2024simtxtseg} converts textual prompts into visual ones via hybrid attention, and TextMatch~\cite{li2024textmatch} leverages text prompts for semi-supervised segmentation. Beyond segmentation, multimodal fusion has been applied to disease detection: MMBCD~\cite{jain2024mmbcd} combines mammograms with clinical history to associate text embeddings with visual regions for breast cancer detection; MEDBind~\cite{gao2024medbind} anchors tri-modal binding of chest X-rays, ECG signals, and textual data; TQx~\cite{nguyen2024towards} aligns pathology images and text using vision–language pre-training to improve quantitative analysis.
Architectures for deeper fusion include TGCAM~\cite{guo2024common}, which uses text-guided common attention with iterative text refinement, and MMI~\cite{bui2024visual}, which matches and fuses image–text features through cross-attention for chest lesion segmentation. Overall, integrating visual and textual modalities yields clear benefits for medical imaging. However, for fine-grained clinical tasks such as laryngopharyngeal cancer T-staging, existing methods struggle to model anatomical relationships and achieve precise cross-modal correspondence between structures and descriptions. To address these issues, we propose an anatomy-aware multimodal framework that jointly models anatomical structure and fine-grained alignment to enhance T-staging performance.

\begin{figure*}[t]
\centering
\includegraphics[width=\linewidth]{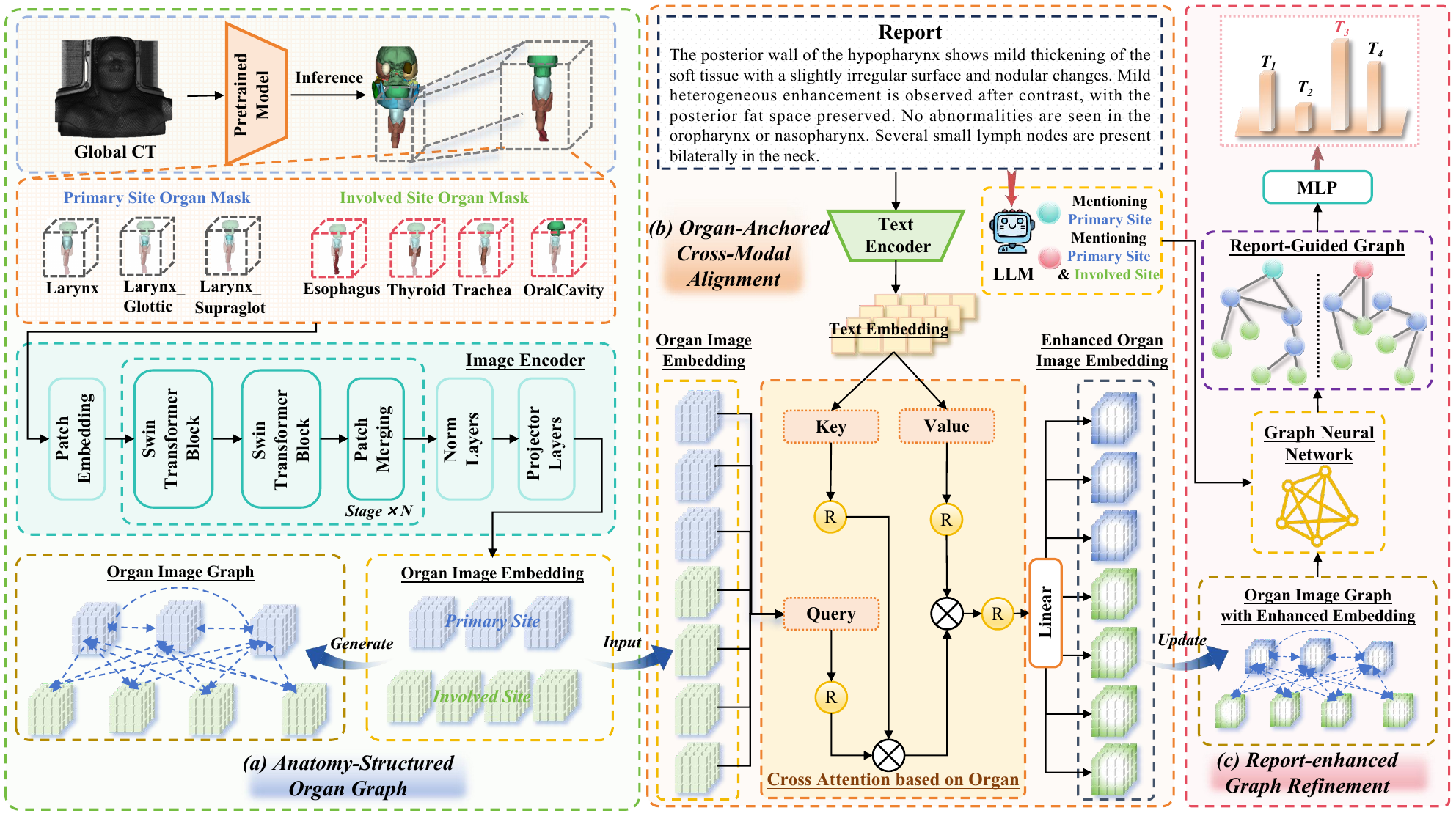}
\caption{
Overview of the proposed framework for laryngopharyngeal cancer T-staging, integrating CT images and radiology reports through three modules: (a) Anatomy-Structured Organ Graph (AOG) for organ-level feature extraction and graph construction, (b) Organ-Anchored Cross-Modal Alignment (OCA) for fine-grained cross-modal alignment, and (c) Report-Enhanced Graph-Refinement (REG) module for structural refinement using clinical invasion information.
}  
\label{fig:framework}
\vspace{-10pt}
\end{figure*}
\section{Methods}
\subsection{Overview}
The proposed anatomy-aware fine-grained multimodal framework for laryngopharyngeal cancer T-staging using CT images and radiology reports is illustrated in Figure~\ref{fig:framework}. Our framework comprises three key components: an \emph{Anatomy-Structured Organ Graph (AOG)}, an \emph{Organ-Anchored Cross-Modal Alignment (OCA)} module, and a \emph{Report-Enhanced Graph-Refinement  (REG)} module. First, AOG addresses the challenge of modeling complex invasion patterns by using pre-trained organ segmentations and anatomy-guided connectivity rules to construct an organ-level graph that encodes potential spread between primary sites and surrounding structures. OCA then performs fine-grained, organ-level image–text interaction, enhancing CT-derived organ features with organ-specific cues from the radiology report while preserving the anatomy-driven topology. Finally, REG uses a large language model (LLM) to incorporate report-derived invasion statements as a guidance signal on top of AOG, refining the multimodal organ graph without discarding anatomy-based risk connections. In the following sections, we describe these three components in detail.

\textbf{Notation:} 
For clarity, we summarize the main symbols and their definitions in Table~\ref{tab:notations}, 
which are consistently used throughout our framework description.

\begin{table}[ht]
\centering
\caption{List of notations used in our framework}
\label{tab:notations}
\begin{tabular}{p{3cm}p{\dimexpr\linewidth-3cm-4\tabcolsep\relax}}
\toprule
\textbf{Symbol} & \textbf{Description} \\ \midrule
$\mathcal{I} = \{I_1, I_2, \dots, I_N\}$ & Set of $N$ CT volumes \\
$I_i \in \mathbb{R}^{H \times W \times D}$ & The $i$-th CT volume with $D$ axial slices of height $H$ and width $W$ \\
$\mathcal{R} = \{R_1, R_2, \dots, R_N\}$ & Corresponding radiology reports \\
$K = 7$ & Number of predefined anatomical structures \\
$\mathcal{P} = \{1, 2, 3\}$ & Indices of primary sites \revB{(supraglottic, glottic, subglottic)} \\
$\mathcal{T} = \{4, 5, 6, 7\}$ & Indices of potentially invaded organs \revB{(esophagus, thyroid gland, trachea, oral cavity)} \\
$d$ & Dimension of feature representations \\
\bottomrule
\end{tabular}
\end{table}

\subsection{Anatomy-Structured Organ Graph}

\noindent\textbf{Motivation.}
Current approaches for laryngopharyngeal cancer T-staging either process whole CT volumes without explicit anatomical constraints or rely on manual tumor segmentation. Both strategies struggle to capture the progressive, anatomy-structured spread of tumor invasion from primary sites to surrounding organs:

\begin{itemize}
\item[\faThumbsODown] \textit{Missing anatomy-structured invasion context:} Operating on global volumes or isolated tumor masks without an explicit organ-level graph underrepresents the relationships between primary sites and anatomically at-risk organs, limiting the ability to reason about stage-defining invasion patterns.
\end{itemize}

\noindent\raisebox{-0.2em}{\includegraphics[scale=0.25]{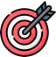}} To address this, we observe that pre-trained organ segmentation models provide reliable anatomical anchors due to the relatively consistent appearance of laryngopharyngeal and surrounding structures. Combined with clinical prior knowledge about typical routes of spread, this motivates constructing an \emph{Anatomy-Structured Organ Graph (AOG)} that explicitly encodes organ-level relationships between primary sites and all anatomically plausible invasion targets, rather than relying solely on tumor masks or reported invasion.

\subsubsection{Organ-specific Feature Extraction}
Effective feature extraction from 3D CT volumes poses significant challenges in laryngopharyngeal cancer staging. The high dimensionality of volumetric data combined with the relatively small size of tumor regions makes it difficult to capture discriminative features directly from whole CT scans. While tumor segmentation seems like a natural solution, the heterogeneous appearance of tumors and the requirement for labor-intensive annotations limit its practical application.

To address these challenges, we leverage organ segmentation as a more reliable intermediate representation. Compared to tumors, anatomical organs exhibit more consistent appearances and clear boundaries, enabling robust feature extraction through pre-trained segmentation models \cite{ye2022comprehensive,guo2025continual}. Specifically, given a CT volume $I_i \in \mathbb{R}^{H \times W \times D}$, we first employ a pre-trained organ segmentation model to obtain masks for all $K$ organs:
\begin{equation}
M_k = \mathcal{S}(I_i), k \in \{1,...,K\}
\end{equation}
where $M_k \in \{0,1\}^{H \times W \times D}$ represents the binary mask for the $k$-th organ, and $\mathcal{S}$ denotes the segmentation model.

To extract organ-specific features, we adopt a Swin Transformer backbone pre-trained on medical images. For each organ mask $M_k$, we obtain its feature representation through:
\begin{equation}
f_k = \mathcal{F}(I_i \odot M_k)
\end{equation}
where $\odot$ denotes element-wise multiplication, and $\mathcal{F}$ represents the feature extraction network that maps each organ region to a feature vector $f_k \in \mathbb{R}^d$. This approach ensures that the extracted features are specifically focused on relevant anatomical structures while leveraging the power of modern vision transformers.

\subsubsection{Anatomy-guided Graph Construction}
After obtaining organ-specific features, a critical challenge is how to effectively model the progressive nature of tumor invasion. Inspired by clinical knowledge that laryngopharyngeal cancer typically spreads from primary sites to surrounding structures in specific patterns, we construct an anatomically-informed graph structure that explicitly captures these invasion relationships.

Formally, we construct a directed graph $\mathcal{G} = (\mathcal{V}, \mathcal{E})$, where the node set $\mathcal{V}$ consists of organ features $\{f_k\}_{k=1}^K$. To effectively capture tumor invasion patterns, we first transform the organ features through a shared feature projection layer:
\begin{equation}
h_k = \sigma(W_p f_k + b_p)
\end{equation}
where $W_p \in \mathbb{R}^{d \times d}$ and $b_p \in \mathbb{R}^{d}$ are learnable parameters, and $\sigma$ denotes the GELU activation function. These projected features $\{h_k\}_{k=1}^K$ serve as node representations in our graph.

The edge set $\mathcal{E}$ is carefully designed to reflect two types of crucial relationships in laryngopharyngeal cancer staging:
\begin{equation}
\mathcal{E} = \mathcal{E}_p \cup \mathcal{E}_t
\end{equation}
where $\mathcal{E}_p$ represents connections between primary sites:
\begin{equation}
\mathcal{E}_p = \{(i,j) | i,j \in \mathcal{P}, i \neq j\}
\end{equation}
and $\mathcal{E}_t$ denotes connections between primary sites and their potential invasion targets:
\begin{equation}
\mathcal{E}_t = \{(i,j) | i \in \mathcal{P}, j \in \mathcal{T}\}
\end{equation}

To enable effective information propagation along these anatomically meaningful pathways, we employ Graph Attention Networks (GAT) for message passing:
\begin{equation}
h_k^{(l+1)} = \sigma(\sum_{j \in \mathcal{N}(k)} \alpha_{kj}^{(l)} W^{(l)} h_j^{(l)})
\end{equation}
where $\mathcal{N}(k)$ denotes the neighbors of node $k$, $\alpha_{kj}^{(l)}$ represents the attention coefficient between nodes $k$ and $j$ at layer $l$, and $W^{(l)} \in \mathbb{R}^{d \times d}$ is the learnable transformation matrix. This attention mechanism allows the model to dynamically focus on the most relevant invasion relationships for staging decisions.

Through this graph-based feature refinement, organ representations are enhanced by incorporating both local anatomical context and global invasion patterns. The resulting node features provide a comprehensive view of potential tumor spread, serving as a foundation for subsequent cross-modal enhancement.

\subsection{Organ-Anchored Cross-Modal Alignment}
\noindent\textbf{Motivation.}
Current approaches for integrating medical images and reports typically rely on global feature fusion, where CT volumes and entire reports are encoded into holistic embeddings and fused at a coarse level. Such strategies overlook the organ-level structure that is present in both modalities:

\begin{itemize}
\item[\faThumbsODown] \textit{Coarse image–text Alignment:} Global fusion methods cannot reliably associate specific anatomical structures with their relevant clinical descriptions in the radiology report, leaving organ features weakly grounded in textual evidence.
\end{itemize}

\noindent\raisebox{-0.2em}{\includegraphics[scale=0.25]{potential.png}} To address this, we let organ nodes from CT serve as queries over the report and propose an \emph{Organ-Anchored Cross-Modal Alignment (OCA)} module that performs fine-grained image–text interaction at the organ level, yielding text-enhanced organ representations that remain consistent with the anatomy-structured organ graph.

\subsubsection{Text Feature Encoding}
Building upon the anatomical topology constructed in Section III.B, we now focus on enhancing organ representations with complementary information from radiology reports. Effective encoding of radiology reports requires careful consideration of medical domain knowledge and terminology. Unlike general text processing tasks, radiology reports contain specialized medical vocabulary and context-dependent descriptions of anatomical structures. To accurately capture this domain-specific information, we employ a pre-trained biomedical language model\revB{, specifically BioBERT~\cite{lee2020biobert} (\texttt{dmis-lab/biobert-base-cased-v1.2})}.

Given a radiology report $R_i$, we first tokenize it into a sequence:
\begin{equation}
R_i \rightarrow \{w_1, w_2, ..., w_L\}
\end{equation}
where $w_l$ represents individual tokens and $L$ is the sequence length. These tokens are then processed through the language model to obtain contextualized representations:

\begin{equation}
T = \mathcal{E}(\{w_1, w_2, ..., w_L\})
\end{equation}
where $T = \{t_1, t_2, ..., t_L\}$, $t_l \in \mathbb{R}^{d}$ represents the hidden state of the $l$-th token, and $\mathcal{E}$ denotes the text encoder. The obtained features are then normalized to ensure stable training while preserving medical semantics.

\subsubsection{Organ-guided Cross-modal Attention}
With organ-specific visual features and textual representations available, a key challenge is how to establish precise correspondence between them. Unlike conventional approaches that apply global feature fusion, we propose an organ-guided attention mechanism where organ features actively query relevant textual descriptions.

Specifically, for each organ $k$, we use its visual feature $h_k$ as a query to attend to the text tokens:
\begin{equation}
\alpha_{k,l} = \text{softmax}(\frac{(W_q h_k)(W_k t_l)^T}{\sqrt{d}})
\end{equation}
where $W_q, W_k \in \mathbb{R}^{d \times d}$ are learnable projection matrices. The enhanced organ features are then computed through weighted aggregation of text features:
\begin{equation}
\tilde{h}_k = h_k + \sum_{l=1}^L \alpha_{k,l}(W_v t_l)
\end{equation}
where $W_v \in \mathbb{R}^{d \times d}$ is the value projection matrix. This design enables each organ to selectively attend to its relevant textual descriptions, facilitating fine-grained cross-modal alignment.

Through this organ-guided attention mechanism, each organ's representation is enhanced with relevant clinical evidence from the report. The enhanced features $\{\tilde{h}_k\}_{k=1}^K$ preserve organ-specific characteristics while incorporating complementary textual information. These text-enhanced organ features then replace the original node representations in our anatomical graph $\mathcal{G}$, enabling subsequent graph-based reasoning to benefit from both structural and textual evidence for accurate staging assessment.

\subsection{Report-Enhanced Graph-Refinement }
\noindent\textbf{Motivation.}
Radiology reports often contain explicit statements about which organs are invaded. However, when report information is used only at the feature level (e.g., to enhance organ embeddings), these discrete invasion findings do not directly influence the structure of the organ graph: all anatomically at-risk organs remain connected in the same way, regardless of whether they are documented as invaded.

\begin{itemize}
\item[\faThumbsODown] \textit{Lack of report-aware structural guidance:} Feature-level fusion alone cannot distinguish organs that are explicitly described as invaded from other anatomically at-risk organs in the graph, so high-confidence invasion evidence from reports is not fully exploited to guide graph-based reasoning.
\end{itemize}

\noindent\raisebox{-0.2em}{\includegraphics[scale=0.25]{potential.png}} 
To address this, we propose a \emph{Report-Enhanced Graph-Refinement  (REG)} module that parses organ-specific invasion evidence from the report and injects it into the anatomy-structured organ graph via a dedicated text node connected only to primary sites and explicitly invaded organs, while preserving the original organ--organ connectivity.

\subsubsection{Text Node Construction}
To fully leverage the structured invasion information in radiology reports, we propose to construct a dedicated text node that serves as a centralized repository of clinical evidence. Unlike previous approaches that treat textual information as auxiliary features, our text node is designed to explicitly encode invasion relationships described in the reports.

Given a radiology report $R_i$, we first employ a large language model to extract structured invasion information through prompt engineering:
\begin{equation}
\mathcal{I} = \text{LLM}(P(R_i))
\end{equation}
where $P(\cdot)$ denotes our designed prompt template that instructs the model to identify invaded organs, and $\mathcal{I} \subseteq \{1,...,K\}$ represents the set of invaded organs mentioned in the report. \textcolor{red}{This is a straightforward information extraction task: the LLM identifies whether specific target organ names (esophagus, thyroid gland, trachea, oral cavity) appear in the context of explicit invasion statements. The main subtlety is distinguishing ``thyroid gland'' (the endocrine organ) from ``thyroid cartilage'' (the laryngeal structure), which is handled through targeted prompt design. Specifically, the prompt is: \textit{``Identify which of the following structures are explicitly described as invaded by the tumor: [esophagus, thyroid gland, trachea, oral cavity]. Only include structures with positive invasion statements; exclude negations. Note: thyroid gland (endocrine organ) and thyroid cartilage (laryngeal structure) are different. Return a JSON list of invaded organs.''} We use Claude 3.5 Haiku~\cite{anthropic2024claude} in our implementation; we verified that GPT-4o~\cite{openai2024gpt4o} and Qwen-2.5-72B~\cite{yang2024qwen25} yield exactly the same outputs as Claude 3.5 Haiku on all 439 reports, and manual review confirmed all extractions are correct.} This structured parsing enables us to capture explicit clinical evidence about tumor spread patterns.

The text node feature is then constructed using the global text representation:
\begin{equation}
h_t = W_t t_1 + b_t
\end{equation}
where $t_1 \in \mathbb{R}^d$ is the [CLS] token representation from the report encoder, and $W_t \in \mathbb{R}^{d \times d}$, $b_t \in \mathbb{R}^d$ are learnable parameters that align the text feature dimension with organ representations.

\subsubsection{Graph Refinement}
With the structured invasion information extracted, we enhance our anatomical graph by incorporating this clinical evidence. The enhanced graph $\hat{\mathcal{G}} = (\hat{\mathcal{V}}, \hat{\mathcal{E}})$ extends the original structure with the text node and report-guided connections:
\begin{equation}
\hat{\mathcal{V}} = \mathcal{V} \cup \{v_t\}
\end{equation}
\begin{equation}
\hat{\mathcal{E}} = \mathcal{E} \cup \{(k,t)| k \in \mathcal{P}\} \cup \{(k,t)| k \in \mathcal{T}, k \in \mathcal{I}\}
\end{equation}
where the text node is connected to all primary sites by default, and additionally linked to target organs that are explicitly mentioned as invaded in the report ($k \in \mathcal{I}$). This design ensures that the graph structure directly reflects clinical observations of invasion patterns.

The final node representations are obtained through message passing on this report-enhanced graph:
\begin{equation}
z_k = \text{GNN}(\{\tilde{h}_k\}_{k=1}^K \cup \{h_t\}, \hat{\mathcal{E}})
\end{equation}
followed by global average pooling for staging classification:
\begin{equation}
y = \text{MLP}(\frac{1}{K+1}\sum_{k \in \hat{\mathcal{V}}} z_k)
\end{equation}

By explicitly incorporating structured invasion information from reports into the graph topology, our approach ensures that clinical evidence directly guides the feature refinement process, leading to more accurate staging decisions.

\subsubsection{Training Strategy}
We train our framework end-to-end using cross-entropy loss for T-staging classification:
\begin{equation}
\mathcal{L} = -\sum_{c=1}^C y_c\log(\hat{y}_c)
\end{equation}
where $y_c$ denotes the ground truth label and $\hat{y}_c$ represents the predicted probability for class $c$. 

\definecolor{lightpink}{RGB}{253, 244, 241}
\definecolor{lightblue}{RGB}{245, 255, 255}
\definecolor{lightgreen}{RGB}{255, 255, 241}
\newcommand{\yes}{{\color{mydarkdarkgreen}\faCheck}}
\newcommand{\no}{\phantom{\faCheck}}

\begin{table*}[t]\small
\captionsetup{font=small}
\caption{{
\textbf{Comparison with Different Approaches for Laryngopharyngeal Cancer T-staging Classification.} We evaluate different methods across binary (Early T1–T2 vs. Late T3–T4) and three-class (T1, T2, T3–T4) classification settings.
}}
\label{tab:compare_sota_vila}
\vspace{-5pt}
\centering
\scriptsize{
\resizebox{\linewidth}{!}{
\setlength\tabcolsep{1.pt}
\renewcommand\arraystretch{1.2}

\begin{tabular}{
  >{\columncolor{mylightgray}}r||
  >{\columncolor{mylightgray}}c>{\columncolor{mylightgray}}c|>{\columncolor{lightblue}}c>{\columncolor{lightgreen}}c |
  >{\columncolor{mylightgray}}c>{\columncolor{mylightgray}}c|>{\columncolor{lightblue}}c>{\columncolor{lightgreen}}c |
  >{\columncolor{mylightgray}}c>{\columncolor{mylightgray}}c
}
\hline\thickhline
\rowcolor{gray!20}
&
\multicolumn{4}{c|}{\textbf{2-Class (Early T1–T2 vs. Late T3–T4}} &
\multicolumn{4}{c|}{\textbf{3-Class (T1, T2, T3–T4)}} &
\multicolumn{2}{c}{\textbf{}}
\\
\cline{2-11}
\rowcolor{gray!20}
& F1 (Internal) & Acc (Internal) & F1 (External) & Acc (External)
& F1 (Internal) & Acc (Internal) & F1 (External) & Acc (External)
& Report & ROI
\\
\hline\hline

\multicolumn{11}{l}{\textbf{\textit{\repft{}}}}\\ 

SwinViT~\cite{hatamizadeh2021swin}
& \tworowdown{58.82}{0.57} & \tworowdown{60.59}{1.47} & \tworowup{43.64}{3.74} & \tworowup{48.57}{3.54}
& \tworowdown{42.02}{2.49} & \tworowdown{42.60}{2.79} & \tworowup{33.10}{3.51} & \tworowup{34.93}{3.15}
& \no & \no
\\

VDPF~\cite{xie2024vdpf}
& \tworowdown{71.47}{3.71} & \tworowdown{73.34}{3.24} & \tworowup{61.99}{3.42} & \tworowup{63.23}{3.42}
& \tworowdown{51.53}{2.72} & \tworowdown{53.98}{1.83} & \tworowup{42.40}{4.27} & \tworowup{52.59}{2.15}
& \no & \yes
\\

MRSN~\cite{sun2025multi}
& \tworowdown{59.68}{4.13} & \tworowdown{61.97}{4.22} & \tworowup{49.10}{3.14} & \tworowup{49.35}{3.16}
& \tworowdown{43.57}{3.31} & \tworowdown{43.97}{3.68} & \tworowup{34.53}{3.59} & \tworowup{39.16}{4.54} 
& \no & \yes
\\

\hdashline
\multicolumn{11}{l}{\textbf{\textit{\selft}}}\\ 

CPAM~\cite{lee2023text}
& \tworowdown{74.83}{3.09} & \tworowdown{76.53}{2.64} & \tworowup{69.99}{5.03} & \tworowup{70.60}{4.43}
& \tworowdown{56.08}{3.97} & \tworowdown{59.68}{3.77} & \tworowup{45.39}{3.67} & \tworowup{52.54}{2.34}
& \no & \yes
\\

TGCAM~\cite{guo2024common}
& \tworowdown{76.04}{3.23} & \tworowdown{77.67}{2.60} & \tworowup{74.09}{3.11} & \tworowup{74.38}{2.90}
& \tworowdown{55.94}{2.14} & \tworowdown{57.77}{3.16} & \tworowup{50.93}{3.97} & \tworowup{55.48}{5.24}
& \yes & \no \\
MMBCD~\cite{jain2024mmbcd}
& \tworowdown{77.35}{3.00} & \tworowdown{77.84}{4.08} & \tworowup{73.54}{4.47} & \tworowup{73.89}{4.09}
& \tworowdown{58.44}{4.84} & \tworowdown{61.14}{4.99} & \tworowup{46.35}{2.89} & \tworowup{56.00}{2.77}
& \yes & \yes \\

LViT~\cite{li2023lvit}
& \tworowdown{75.54}{2.39} & \tworowdown{76.59}{2.69} & \tworowup{68.83}{4.44} & \tworowup{70.14}{3.47}
& \tworowdown{55.92}{5.50} & \tworowdown{57.85}{5.24} & \tworowup{49.12}{10.21} & \tworowup{53.83}{11.08}
& \yes & \no \\

MMI~\cite{bui2024visual}
& \tworowdown{77.59}{4.01} & \tworowdown{78.57}{4.42} & \tworowup{69.88}{5.11} & \tworowup{71.15}{4.09}
& \tworowdown{57.26}{1.57} & \tworowdown{58.99}{2.25} & \tworowup{51.54}{4.74} & \tworowup{54.75}{5.78} 
& \yes & \no \\

Ours
&  \tworowdown{\textbf{80.83}}{5.35} & \tworowdown{\textbf{81.53}}{5.41} & \tworowup{\textbf{76.46}}{1.35} & \tworowup{\textbf{76.50}}{1.34}
& \tworowdown{\textbf{61.43}}{3.25} & \tworowdown{\textbf{63.33}}{3.25} & \tworowup{\textbf{60.27}}{3.03} & \tworowup{\textbf{63.78}}{2.98}
& \yes & \yes
\\
\hline\thickhline
\end{tabular}}}
\end{table*}

\section{Experiments}
\subsection{Dataset}
To the best of our knowledge, we collect the first and largest dataset for CT-based laryngopharyngeal cancer T-staging, comprising 656 patients across four medical centers. More importantly, all cases are paired with CT reports for multimodal T-staging assessment. Our dataset substantially exceeds the scale of previous studies: \revB{prior laryngopharyngeal and head-and-neck staging studies are single-center and considerably smaller (118--319 patients~\cite{liu2024machine,ren2018magnetic,wang2019radiomic,chen2023combined}) and rely on imaging alone without paired radiology reports.}
\subsubsection{Dataset Composition and Evaluation Protocol}
Our dataset consists of four cohorts: (1) Fudan University Zhongshan Hospital (FUZH) as the primary cohort with 439 patients, each with contrast-enhanced CT scans and corresponding radiology reports; (2) Chang Gung Memorial Hospital (CGMH) with 25 cases; (3) Eye \& ENT Hospital of Fudan University (EENT) with 87 cases; and (4) The Cancer Genome Atlas (TCGA) with 105 cases for evaluating generalizability on public data~\cite{TCIA2013}. All cases are pathologically confirmed with detailed T-staging distributions as shown in Table~\ref{tab:dataset_statistics}.
For TCGA data~\cite{TCIA2013}, which included only CT scans and pathological-confirmed T-stages, we manually annotated corresponding CT reports following standard radiological assessment protocols.
\revB{Specifically, these CT reports were written by a board-certified radiologist and then reviewed and verified by a second, with any disagreements resolved by consensus; the radiologists described only imaging findings and did not assign or infer the T-stage.}
FUZH is used for 5-fold cross-validation, while CGMH, EENT, and TCGA~\cite{TCIA2013} form the external test set.

\newcolumntype{Y}{>{\centering\arraybackslash}X}
\begin{table}[t]\small
\centering
\caption{Dataset statistics across four medical centers.}
\label{tab:dataset_statistics}
\resizebox{\linewidth}{!}{
\setlength\tabcolsep{3.5pt}
\begin{tabular}{lccccc|ccc}
\hline\thickhline
\rowcolor{mygray}
\textbf{Center} & \textbf{T1} & \textbf{T2} & \textbf{T3} & \textbf{T4} & \textbf{Total} & \textbf{Split} & \textbf{Cases} & \textbf{\%} \\
\hline\hline
FUZH & 151 & 116 & 99 & 73 & 439 & Training & 439 & 66.9\% \\
\hdashline
CGMH & 2 & 4 & 4 & 15 & 25 & \multirow{3}{*}{Test} & \multirow{3}{*}{217} & \multirow{3}{*}{33.1\%} \\
EENT & 20 & 29 & 17 & 21 & 87 & & & \\
TCGA~\cite{TCIA2013} & 24 & 27 & 18 & 36 & 105 & & & \\
\hline
\rowcolor[HTML]{D7F6FF}
\textbf{Total} & \textbf{197} & \textbf{176} & \textbf{138} & \textbf{145} & \textbf{656} & \textbf{All} & \textbf{656} & \textbf{100\%} \\
\hline\thickhline
\end{tabular}
}
\end{table}

\textcolor{red}{Importantly, the ground-truth T-stage labels in our dataset are derived from pathological confirmation following AJCC 8th edition criteria, not from CT reports. None of the 439 training reports contain any T-stage designation --- all reports describe imaging findings following standard clinical templates.}

We consider two clinically relevant classification settings: (1) Binary classification between early stages (T1--T2) and late stages (T3--T4), reflecting fundamentally different treatment strategies---patients with early-stages receive organ-preserving procedures while advanced cases require aggressive interventions; (2) Three-class classification: T1 vs. T2 vs. T3--T4, where T1 and T2 are distinguished due to different tumor infiltration patterns and recurrence risks, while T3 and T4 are merged considering similar treatment protocols. 

\begin{table}[t]\small
\centering
\caption{\small\textbf{Ablation Study of Key Components.} Performance comparison of different component combinations for binary and three-class T-staging tasks on a representative internal fold, reported as mean $\pm$ standard deviation over three random seeds. AOG, OCA, and REG represent Anatomy-Structured Organ Graph, Organ-Anchored Cross-Modal Alignment, and Report-Enhanced Graph-Refinement, respectively.}
\label{tab:component}
\resizebox{\linewidth}{!}{
\setlength\tabcolsep{3.8pt}
\begin{tabular}{ccc||cc|cc}
\hline \thickhline
\rowcolor{mygray}
&&& \multicolumn{2}{c}{Two Class} & \multicolumn{2}{c}{Three Class} \\
\cline{4-5}\cline{6-7}
\rowcolor{mygray}
\multirow{-2}{*}{AOG} &
\multirow{-2}{*}{OCA} &
\multirow{-2}{*}{\textsc{REG}} &
F1 (macro) & Acc & F1 (macro) & Acc \\
\hline\hline
& & & 58.76 $\pm$ 2.68 & 61.36 $\pm$ 2.27 & 39.16 $\pm$ 1.60 & 40.15 $\pm$ 1.74\\
\ding{51} & & & 77.96 $\pm$ 1.23 & 78.41 $\pm$ 1.14 & 52.80 $\pm$ 2.31 & 53.41 $\pm$ 2.27\\
& \ding{51} & & 80.97 $\pm$ 1.28 & 82.20 $\pm$ 1.31 & 58.47 $\pm$ 1.06 & 59.09 $\pm$ 1.14\\
\ding{51}& \ding{51} & & 81.63 $\pm$ 1.71 & 83.71 $\pm$ 1.74 & 61.52 $\pm$ 1.04 & 64.02 $\pm$ 1.31\\
\rowcolor[HTML]{D7F6FF}
\ding{51} & \ding{51} & \ding{51} & 83.96 $\pm$ 0.50 & 84.85 $\pm$ 0.66 & 64.53 $\pm$ 0.70 & 65.53 $\pm$ 0.66\\
\hline \thickhline
\end{tabular}
}

\end{table}

\subsection{Experimental Setup}
\subsubsection{Implementation Details}
All experiments are conducted using PyTorch~\cite{paszke2019pytorch} on an NVIDIA H20 GPU. For data preprocessing, all CT volumes are resampled to an isotropic spacing of 1mm$^3$ and resized to a unified spatial dimension of 128$\times$128$\times$128. The intensity values are normalized to the range [0, 1], with Hounsfield Unit normalization applied using a window center of 50 and a window width of 450 to enhance the visibility of soft tissues in clinical settings. We train our model with a batch size of 8 using the Adam optimizer. The learning rate starts at \(1\times10^{-6}\) and decays to \(1\times10^{-7}\) following a cosine decay schedule, with a weight decay of \(1\times10^{-5}\).
\revB{Organ masks for all four cohorts are produced by the same publicly available pre-trained segmentation model (Continual Segment~\cite{ji2023continual}, a single unified model covering 143 whole-body organs in CT), applied without any site-specific fine-tuning or adaptation.}

\subsubsection{Evaluation Criteria}
We evaluate model performance using Accuracy and Macro F1-score. Accuracy reflects the proportion of correct predictions, F1-score balances precision 
and recall, and the macro-averaged F1-score treats all classes equally by computing the 
unweighted mean of class-wise F1-scores. 

\subsection{Comparison to State-of-the-Arts}
We compare the proposed method with image-based classification approaches, such as SwinViT~\cite{hatamizadeh2021swin}, VDPF~\cite{xie2024vdpf}, MRSN~\cite{sun2025multi}, and several state-of-the-art multimodal learning methods (CPAM~\cite{lee2023text}, TGCAM~\cite{guo2024common}, MMBCD~\cite{jain2024mmbcd}, LViT~\cite{li2023lvit}, MMI~\cite{bui2024visual}). The results are summarized in Table~\ref{tab:compare_sota_vila}.
Our method demonstrates superior performance across all experimental settings. In the binary classification task (Early T1–T2 vs. Late T3–T4), our method achieves the highest performance on both internal and external test sets. For internal testing, we achieve an F1 score of 80.83\% and accuracy of 81.53\%, surpassing the second-best method MMI~\cite{bui2024visual} (77.59\% F1 score and 78.57\% accuracy). On the external test set, our method maintains its leading position with an F1 score of 76.46\% and accuracy of 76.50\%, outperforming TGCAM~\cite{guo2024common} (74.09\% F1 score and 74.38\% accuracy) by a significant margin.
For the more challenging three-class classification task (T1, T2, T3–T4), our method also achieves consistent superior performance. On the internal test set, we achieve an F1 score of 61.43\% and accuracy of 63.33\%, showing notable improvements over MMBCD~\cite{jain2024mmbcd} (58.44\% F1 score and 61.14\% accuracy). The advantage is even more prominent on the external test set, where our method achieves an F1 score of 60.27\% and accuracy of 63.78\%, significantly outperforming the second best method by a large margin of 8.73\% F1 score and 7.78\% accuracy.
\subsection{Ablation Study}

\subsubsection{Effectiveness of Individual Components}
We conduct comprehensive ablation studies to evaluate the contribution of each key component in our framework. As shown in Table~\ref{tab:component}, we systematically analyze different combinations of the three
main modules: Anatomy-Structured Organ Graph (AOG), Organ-Anchored Cross-Modal Alignment (OCA), and Report-Enhanced Graph-Refinement (REG). The baseline model, which directly processes the whole CT volume for prediction
without any anatomical guidance or textual information, achieves only 58.76\% and 39.16\% macro F1-scores in binary and three-class settings, respectively. Adding AOG substantially
improves the performance (77.96\% and 52.80\% F1-scores), demonstrating that explicitly modeling organ-level relationships helps capture tumor invasion patterns. When applying the OCA
module alone, the performance further increases to 80.97\% and 58.47\% for binary and three-class settings, which validates our
hypothesis that fine-grained organ–text alignment provides valuable diagnostic evidence from radiology reports. When combining AOG and OCA, we observe further performance gains (81.63\% and 61.52\% F1-scores), suggesting that
anatomical structure and textual evidence complement each other. Finally, incorporating REG on top of both AOG and OCA yields the best overall performance, with macro F1-scores of
83.96\% and 64.53\% in binary and three-class settings, indicating that the structured parsing of invasion information from reports further enhances the graph-based representation.
These results comprehensively validate the effectiveness of our proposed modules in improving T-staging accuracy.
\revB{To quantify the uncertainty of these comparisons, each ablation configuration is run over three random seeds and reported as mean $\pm$ standard deviation (Table~\ref{tab:component}). On the binary task the macro-F1 is AOG 77.96 $\pm$ 1.23, OCA 80.97 $\pm$ 1.28, AOG+OCA 81.63 $\pm$ 1.71, and the full model 83.96 $\pm$ 0.50; the full model attains both the highest mean and the smallest variance, and REG yields a further improvement over AOG+OCA.}

\begin{table}[t]\small
\centering
\caption{\small\textbf{Comparison of different attention mechanisms} in our OCA on a representative internal fold. The results demonstrate the superiority of our proposed method across both datasets.}
\label{tab:otim_comparison}
\resizebox{\linewidth}{!}{
\setlength\tabcolsep{3.8pt}
\begin{tabular}{c||cc|cc}
\hline \thickhline
\rowcolor{mygray}
& \multicolumn{2}{c}{Internal Test} & \multicolumn{2}{c}{External Test} \\
\cline{2-3}\cline{4-5}
\rowcolor{mygray}
\multirow{-2}{*}{Method} &
F1 (macro) & Acc & F1 (macro) & Acc \\
\hline\hline
Global + Attention & 75.45 & 77.27 & 68.47 & 68.66\\
Local + Attention & 79.04 & 81.81 & 71.16 & 71.76\\
Common Attention & 78.65 & 79.54 & 73.06 & 73.61\\
MultiToken Attention & 69.88 & 71.59 & 68.24 & 68.66\\\rowcolor[HTML]{D7F6FF}
Ours & 81.93 & 82.95 & 76.10 & 76.50\\
\hline \thickhline
\end{tabular}
}

\label{tab:ablation}
\end{table}

\begin{table}[t]\small
\centering
\caption{\small\textbf{Ablation study of anatomical topology modeling.} Comparison between different approaches for modeling anatomical relationships on a representative internal fold. Fully-Connected Graph represents a complete graph connection strategy, Direct Fusion uses concatenated organ masks, and Global Context applies global feature pooling.}
\resizebox{\linewidth}{!}{
\setlength\tabcolsep{3.8pt}
\begin{tabular}{c||cc|cc}
\hline \thickhline
\rowcolor{mygray}
& \multicolumn{2}{c}{Internal Test} & \multicolumn{2}{c}{External Test} \\
\cline{2-3}\cline{4-5}
\rowcolor{mygray}
\multirow{-2}{*}{Method} &
F1 (macro) & Acc & F1 (macro) & Acc \\
\hline\hline

Fully-Connected Graph & 73.91 & 75.00 & 69.00 & 69.12\\
Direct Fusion & 53.94 & 56.82 & 56.10 & 56.68\\
Global Context & 58.26 & 61.36 & 47.02 & 52.07\\
\rowcolor[HTML]{D7F6FF}
Ours (AOG) & 77.76 & 78.41 & 73.33 & 75.00\\
\hline \thickhline
\end{tabular}
}

\label{tab:atm_ablation}
\end{table}

\subsubsection{Impact of Text Encoder Freezing}
\begin{figure}[!t]
\centerline{\includegraphics[width=\columnwidth]{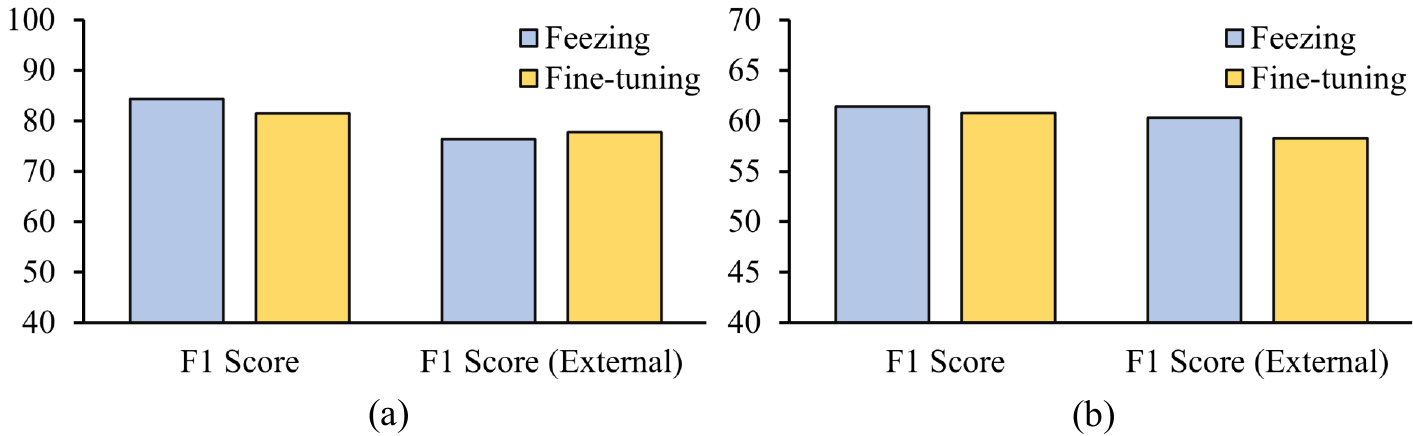}}
\caption{\textbf{Impact of Text Encoder Freezing.} Comparison of model performance with frozen and fine-tuned text encoder on both internal validation and external testing sets. Results are shown for both binary and three-class T-staging tasks.}

\label{freeze}
\end{figure}

We investigate the influence of text encoder fine-tuning strategies on model performance. As shown in Figure~\ref{freeze}, we compare two settings: freezing the text encoder parameters and fine-tuning them during training. For the binary classification task, the fine-tuned encoder achieves slightly better performance on both internal validation (81.46\% vs. 80.83\% F1-score) and external testing (77.77\% vs. 76.46\% F1-score) sets. Similarly, in the three-class setting, both strategies yield comparable results, with frozen encoder showing marginal advantages (61.43\% vs. 60.79\% on internal validation, 60.27\% vs. 58.33\% on external testing). The minimal performance difference between frozen and fine-tuned settings suggests that the pre-trained text encoder already captures robust medical language representations. This observation is particularly valuable from a practical perspective, as freezing the text encoder not only reduces computational costs but also helps prevent potential overfitting on our specific dataset. Therefore, we adopt the frozen text encoder strategy in our final framework.

\subsubsection{Effectiveness of Individual Invaded Organs}
\begin{figure}[!t]
\centerline{\includegraphics[width=\columnwidth]{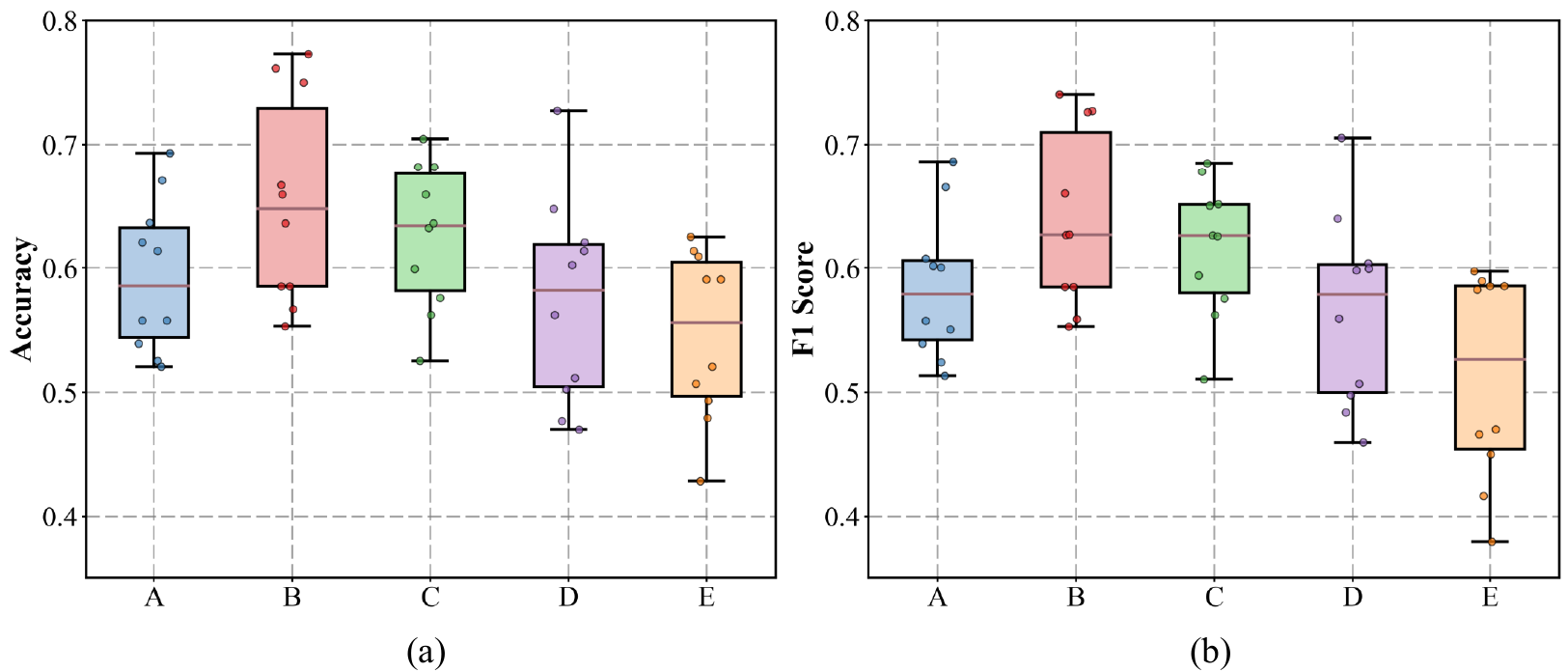}}
\caption{\textbf{Performance Analysis of Individual Invaded Organs.} Box plots showing the distribution of (a) F1-scores and (b) Accuracy for different prediction strategies across five-fold cross-validation and external testing. Comparison includes individual invaded organs (A: Esophagus, B: Thyroid, C: Trachea, D: Oral Cavity) and global CT volume processing (E).}

\label{fig:org}
\end{figure}

To validate the importance of invaded organs in T-staging prediction, we conduct experiments comparing the performance of models trained on individual invaded organs against global CT volume processing. As shown in Figure~\ref{fig:org}, surprisingly, models focusing on single invaded organs often outperform the global approach. Particularly, the thyroid-focused model demonstrates the strongest performance, achieving average F1-scores of 0.716 (internal) and 0.582 (external), significantly surpassing the global baseline's 0.588 (internal) and 0.436 (external). Models focusing on the trachea and esophagus also show consistent advantages over global processing, while the oral cavity model exhibits comparable performance. These results are noteworthy because these invaded organs may not contain the primary tumor site, yet they provide reliable staging predictions. This phenomenon can be attributed to two factors: (1) The invasion status of these organs serves as a strong indicator for T-staging, as specific organ involvement directly correlates with staging criteria (e.g., thyroid invasion indicating advanced stage disease). (2) Compared to processing the entire CT volume, focusing on individual organs provides more precise anatomical context, reducing the interference from irrelevant regions. These findings not only validate our anatomically-informed approach but also suggest that targeted analysis of key invaded organs could be a promising direction for efficient T-staging.

\begin{figure}[!t]
\centerline{\includegraphics[width=\columnwidth]{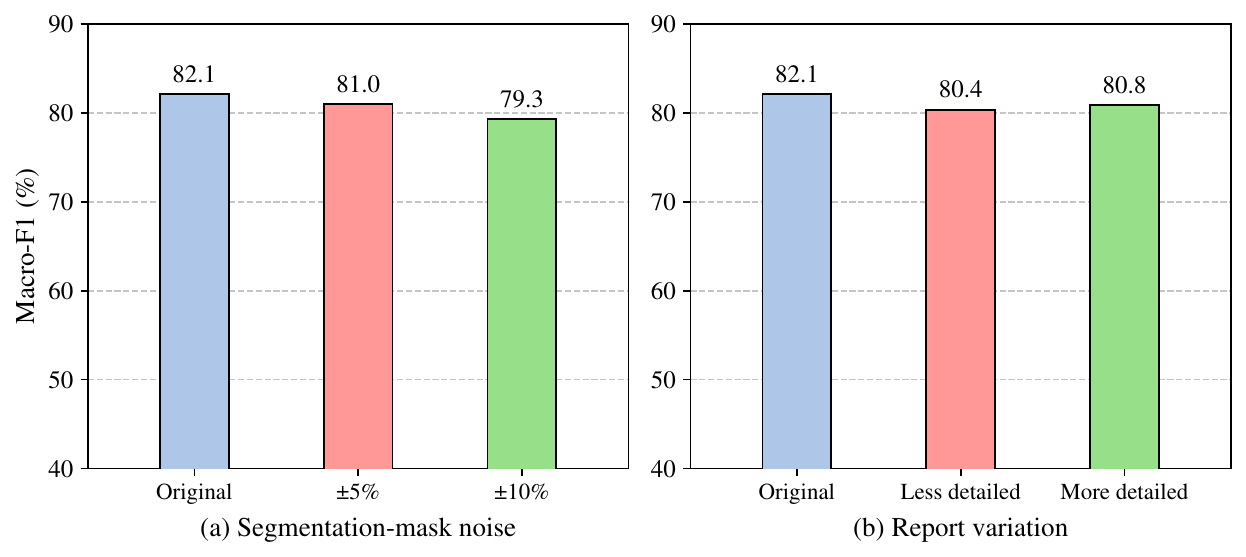}}
\caption{\small\revB{\textbf{Robustness to (a) organ segmentation noise and (b) radiology report variation,} evaluated on a representative internal fold. (a)~Macro-F1 under random organ-mask perturbation at $\pm$5\% and $\pm$10\% boundary shifts; (b)~macro-F1 when each radiology report is rewritten to be less- or more-detailed while the CT images and the trained model are held fixed.}}
\label{fig:robustness}
\end{figure}

\textcolor{red}{\subsection{Robustness to Organ Segmentation Noise \revB{and Report Variation}}}
\textcolor{red}{Since our framework relies on pre-trained organ segmentation to define cropping regions, we evaluate its sensitivity to segmentation errors through a mask perturbation experiment. Starting from the existing auto-segmentation outputs (not ground-truth masks), we apply additional random perturbations to each organ's mask independently: the perturbation direction (dilation or erosion) is randomly chosen per organ, and the magnitude is proportional to the organ's size so that small organs receive smaller absolute perturbations and large organs receive larger ones. As shown in Fig.~\ref{fig:robustness}(a), the F1 score decreases by only 1.1\% and 2.8\% under 5\% and 10\% perturbation levels respectively, confirming that our region-based cropping design is robust to realistic segmentation errors. Since the baseline itself already contains segmentation errors from the automated model, the perturbed conditions represent compounded noise levels beyond what would occur in practice.}

\revB{\noindent\textbf{Robustness to report detail and completeness.} Beyond segmentation noise, we assess the framework's sensitivity to variation in how radiology reports are written. We rewrite each report with a large language model (Claude Haiku), keeping every medical term and measurement verbatim and altering only the level of detail: a less-detailed version drops non-diagnostic content (scan parameters, boilerplate, and routine non-invasion negative statements) while retaining the primary subsite, tumor size, and all positive and negative invasion statements, and a more-detailed version restates the same findings in more elaborate clinical language without adding new information, reflecting the terse and verbose reporting styles seen across radiologists in practice. The same CT images and the frozen model are held fixed throughout. As shown in Fig.~\ref{fig:robustness}(b), macro-F1 stays within 80.4--82.1\% and accuracy within 81.8--83.0\% across the original and both perturbed conditions (at most a 1.7-point change in macro-F1). This indicates that the framework is resilient to the level of detail and completeness of the input report.}

\subsection{Superiority of AOG}
To validate the effectiveness of our Anatomy-Structured Organ Graph (AOG), we conduct ablation studies comparing different approaches for modeling anatomical relationships. As shown in Table \ref{tab:atm_ablation}, we compare our method with three baseline approaches: 1) Fully-Connected Graph, which models organ relationships using a complete graph where all nodes are directly connected; 2) Direct Fusion, which simply concatenates all relevant organ masks as input; and 3) Global Context, which uses global feature pooling to capture contextual information. The results demonstrate the clear advantages of our AOG approach. Specifically, our method achieves significant improvements over all baselines, with F1 scores of 77.76\% and 73.33\% on internal and external tests respectively. This represents improvements of 3.85\% and 4.33\% over the strongest baseline (Fully-Connected Graph) in F1 score. The performance gap is even larger when compared to Direct Fusion and Global Context approaches, showing gains of over 15\% in some metrics. These results validate our design choice in AOG that selectively models anatomical relationships through learned attention mechanisms rather than using either oversimplified global context or potentially noisy fully-connected connections. The consistent superior performance on both internal and external tests also demonstrates the robustness and generalizability of our approach in capturing meaningful anatomical topology information.

\subsection{Superiority of OCA}
\begin{figure*}[t]
\centering
\textcolor{red}{\includegraphics[width=\linewidth]{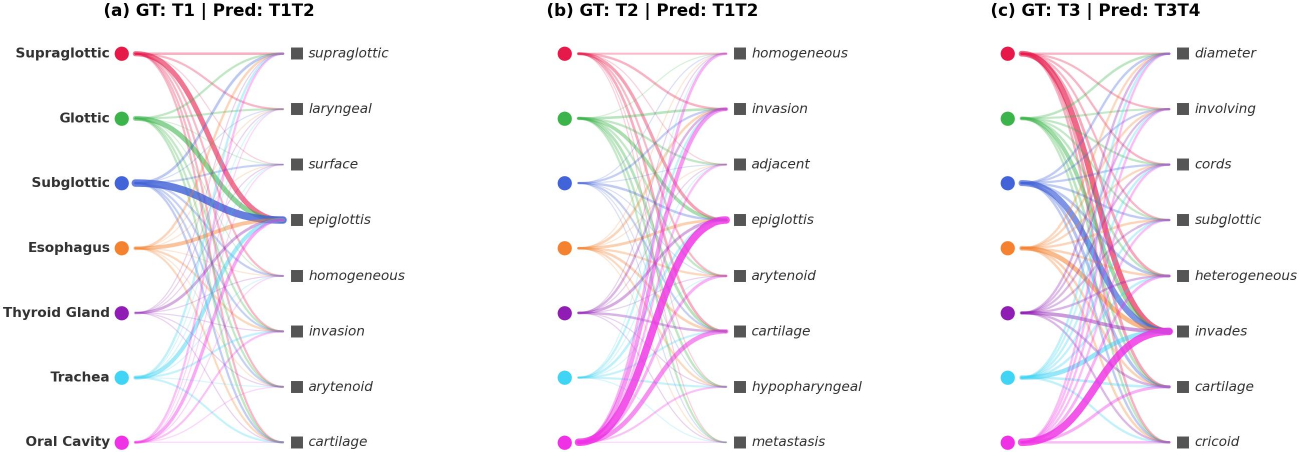}
\caption{\textbf{Visualization of OCA cross-attention weights} for three cases of different T-stages. Each bipartite graph shows organ nodes (left) connected to the most-attended report keywords (right), with line width proportional to attention weight.}
\label{fig:attention_vis}}
\end{figure*}

To validate the effectiveness of our Organ-Anchored Cross-Modal Alignment (OCA), we conduct comprehensive comparisons with several attention-based variants on both internal and external test sets. As shown in Table \ref{tab:otim_comparison}, our proposed method consistently outperforms other attention mechanisms across both settings. For internal testing, our method achieves the highest F1 score of 81.93\% and accuracy of 82.95\%, surpassing the Global+Attention baseline by 6.48\% and 5.68\% respectively. More importantly, on the external test set, our method maintains strong performance with 76.10\% F1 score and 76.50\% accuracy, demonstrating a significant improvement of 7.63\% and 7.84\% over the baseline. The superior performance on both internal and external tests validates that our OCA can effectively capture the fine-grained interactions between organ features and text descriptions, while maintaining good generalization ability. \textcolor{red}{Fig.~\ref{fig:attention_vis} illustrates OCA cross-attention weights for three cases of different T-stages: the early-stage case (T1) attends primarily to anatomical location terms (``supraglottic,'' ``epiglottis''); the T2 case shows cross-organ attention as the Subglottic node attends to adjacent structure tokens; and the advanced case (T3) concentrates on invasion-related keywords (``invades,'' ``cartilage''). These visualizations offer interpretable evidence for how OCA utilizes report text at the organ level.}

\begin{figure*}[t]
\centering
\includegraphics[width=\linewidth]{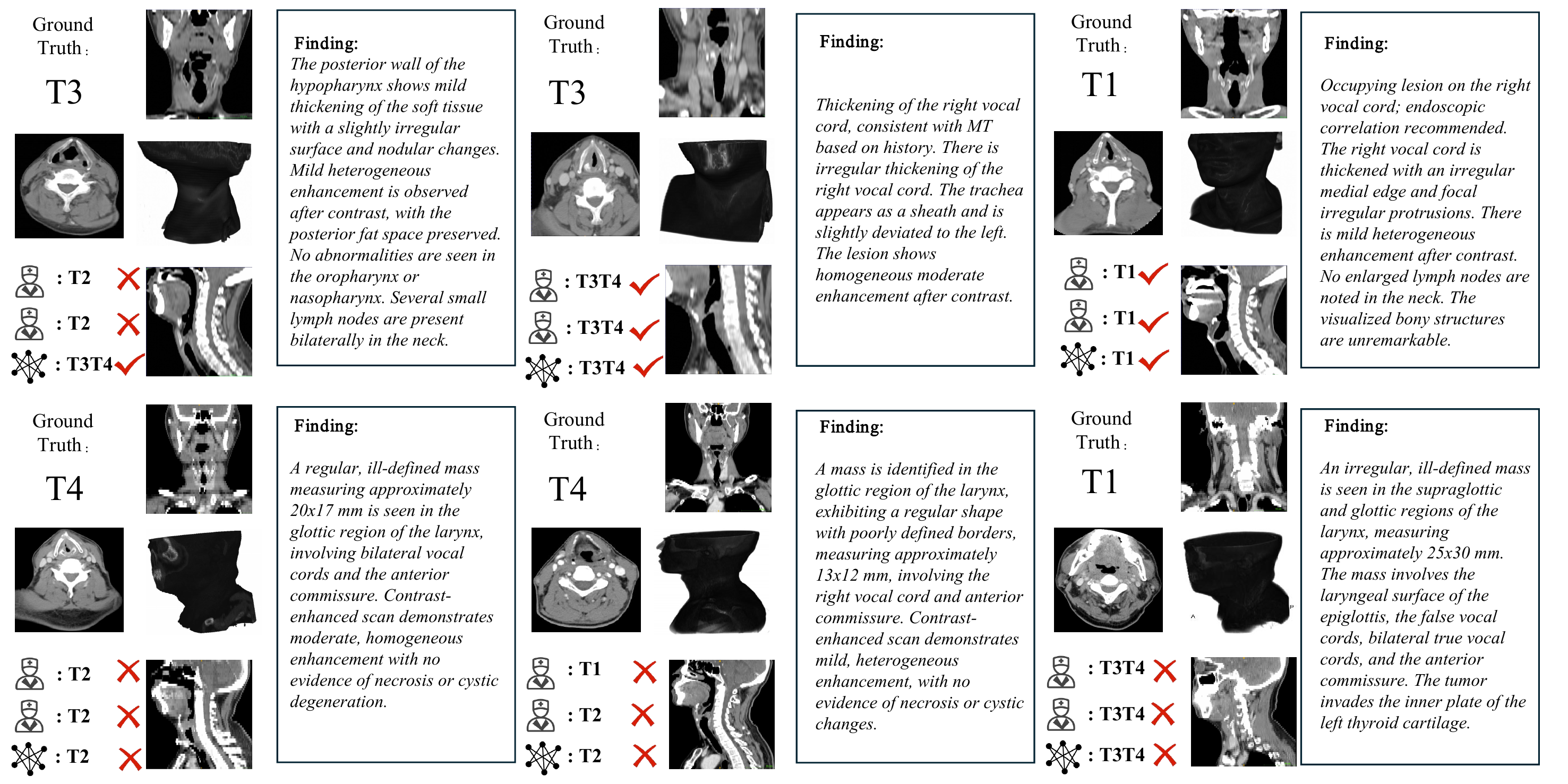}
\caption{
\textcolor{red}{\textbf{Representative correct predictions (top) and failure cases (bottom).} Each case shows the ground truth, radiologist assessments, and our model's prediction. The shared errors in the bottom row reflect inherent limitations of CT-based invasion assessment, detailed in Section~IV-H.}
}
\label{fig:case_study}
\end{figure*}

\textcolor{red}{\subsection{Clinical Validation}}

\textcolor{red}{\subsubsection{Comparison with Radiologists}}
Table~\ref{tab:radio} shows the comparison results of our method and two radiologists on an external test set. For this evaluation, we randomly selected 20 samples from each class, comprising 80 samples in total for the comparison. The proposed method demonstrates competitive performance with the radiologists in both experimental settings. In the Early vs Late classification, our method achieves an F1 score of 76.25\% and accuracy of 76.25\%, performing comparably to Radiologists 1 and 2, who achieve F1 scores of 76.16\% and 74.98\%, and accuracies of 76.25\% and 75.00\%, respectively. In the T1, T2, Late (T3--T4) classification, our method achieves an F1 score of 59.71\% and an accuracy of 62.50\%, outperforming both radiologists by at least 6.08\% in F1 score and 2.50\% in accuracy. These results confirm the effectiveness of our method in predicting laryngopharyngeal cancer T-staging, demonstrating performance on par with or exceeding experienced radiologists. The three representative cases in Fig.~\ref{fig:case_study} demonstrate that, despite potential errors by experienced clinicians, our model correctly performs the T-staging task, showing strong precision and robustness.

\begin{table}[t]\small
\centering
\caption{\small \textbf{Comparison of T-staging prediction performance:} our method versus radiologist assessments.}
\label{tab:radio}
\resizebox{\linewidth}{!}{
\setlength\tabcolsep{3.8pt}
\begin{tabular}{c||cc|cc}
\hline \thickhline
\rowcolor{mygray}
& \multicolumn{2}{c}{Two Class} & \multicolumn{2}{c}{Three Class} \\
\cline{2-3}\cline{4-5}
\rowcolor{mygray}
\multirow{-2}{*}{Method} &
F1 (macro) & Acc & F1 (macro) & Acc \\
\hline\hline
Radiologist 1 & 76.16 & 76.25 & 53.03 & 60.00  \\
Radiologist 2 & 74.98 & 75.00 & 53.63 & 60.00  \\
\rowcolor[HTML]{D7F6FF}
\textbf{Ours} & \textbf{76.25} & \textbf{76.25} & \textbf{59.71} & \textbf{62.50} \\
\hline \thickhline
\end{tabular}
}

\end{table}

\textcolor{red}{\subsubsection{Failure Mode Analysis}}
\textcolor{red}{We analyzed the 16 misclassified cases (out of 88 test samples) in Fold~1 and identified two dominant failure patterns inherently linked to the limitations of CT-based invasion assessment. Over-staging (9/16 cases, 56\%): the model predicted T3--T4 for pathologically confirmed T1--T2 cases. In 7 of these 9 cases, the radiology report explicitly described cartilage invasion that was not confirmed by pathological examination, reflecting CT's known limitation in distinguishing tumor contact from true cartilage invasion~\cite{hao2024deep,guo2020ct}. Under-staging (7/16 cases, 44\%): the model predicted T1--T2 for pathologically confirmed T3--T4 cases. In 5 of these 7 cases, the report contained no invasion description or explicitly stated ``no evidence of invasion,'' despite pathological confirmation. These complementary failure modes highlight an inherent challenge: CT-based assessment can both over-estimate and under-estimate invasion relative to pathological ground truth. Three representative failure cases are shown in Fig.~\ref{fig:case_study}.}

\revB{\noindent\textbf{Cross-modal alignment analysis of failures.} We further examine the misclassified cases in the shared representation space, quantifying the OCA cross-attention mass that the potentially-invaded organ nodes (esophagus, thyroid gland, trachea, oral cavity) place on invasion-related report tokens (Table~\ref{tab:align}). On correctly classified cases the alignment behaves as intended, with low values for true early-stage (0.058) and high values for true advanced-stage cases (0.105). Notably, over-staged cases do exhibit an elevated invasion-token alignment (0.088); this arises because the corresponding reports genuinely contain invasion descriptions that are CT-derived false positives (cf.\ the cartilage discordance in Table~\ref{tab:invasion}), which the cross-modal fusion faithfully incorporates, rather than over-associating invasion with unrelated features. In under-staged cases the alignment is as low as in true early-stage cases (0.058), consistent with these being CT false negatives in which the report records no positive invasion finding and the CT shows none, so no genuine invasion evidence is available for the organ features to bind to. These errors arise from discordance between the available CT/report evidence and the pathological ground truth, which the fusion transmits faithfully rather than mis-aligning. We acknowledge that, because the fusion integrates report evidence, an unreliable invasion description can occasionally tip a borderline case in this misclassified subset. This effect is bounded, however: the prediction is grounded in imaging and enhanced by the report rather than dictated by it, with the image-only AOG variant already attaining 77.96\% F1 (Table~\ref{tab:component}) and the report refining this to 83.96\%. Future work could mitigate these failures through uncertainty-aware fusion that down-weights unreliable invasion descriptions, uncertainty-aware prediction that flags borderline cases for clinical review, and multi-center training to improve cross-institutional generalizability.}

\begin{table}[t]\small
\centering
\caption{\small\revB{\textbf{Cross-modal alignment of correct and misclassified cases.} Mean invasion-token attention mass placed by the potentially-invaded organ nodes on invasion-related report tokens, on the Fold~1 internal test set.}}
\label{tab:align}
\setlength\tabcolsep{8pt}
\begin{tabular}{l||c}
\hline \thickhline
\rowcolor{mygray}
Group & Invasion-token attention \\
\hline\hline
Correctly staged (T1--T2) & 0.058 \\
Correctly staged (T3--T4) & 0.105 \\
Over-staged (T1--T2 $\rightarrow$ T3--T4) & 0.088 \\
Under-staged (T3--T4 $\rightarrow$ T1--T2) & 0.058 \\
\hline \thickhline
\end{tabular}
\end{table}

\textcolor{red}{\section{Discussion}}

\noindent\textcolor{red}{\textbf{Robustness to Organ Segmentation.}}
\textcolor{red}{Head-and-neck organ segmentation is a relatively mature task, with Dice scores of 80--95\% routinely achievable for most anatomical structures~\cite{ye2022comprehensive,ji2023continual}. Moreover, our method uses organ masks only to locate cropping regions, so small boundary errors merely shift the crop window without excluding organ content. The mask perturbation experiment (Fig.~\ref{fig:robustness}(a)) confirms this robustness, with only 2.8\% F1 degradation under $\pm$10\% proportional boundary noise.}

\noindent\textcolor{red}{\textbf{Report--T-stage Independence.}}
\textcolor{red}{Under AJCC 8th edition criteria, thyroid cartilage invasion is a hallmark of advanced T-stage in laryngopharyngeal cancer. To examine whether invasion descriptions in reports could implicitly encode T-stage, we analyzed thyroid cartilage-related descriptions across the 439 FUZH reports (Table~\ref{tab:invasion}). Of 147 reports with positive invasion descriptions, 45 (30.6\%) are from pathologically confirmed early-stage (T1--T2) patients; conversely, among 172 advanced-stage (T3--T4) patients, 70 (40.7\%) lack any positive description. This is because invasion descriptions reflect CT-visible findings --- an imaging observation rather than a pathological diagnosis --- and CT alone cannot determine whether true cartilage invasion has occurred~\cite{hao2024deep,guo2020ct,gilbert2010staging}. This substantial discordance confirms that invasion descriptions do not encode T-stage, ruling out information circularity in our framework.}

\begin{table}[t]\small
\centering
\caption{\small\textcolor{red}{\textbf{Thyroid cartilage invasion descriptions vs.\ pathological T-stage} across 439 FUZH reports. ``Positive invasion'' includes both definite and suspected cartilage invasion descriptions in CT reports.}}
\label{tab:invasion}
\resizebox{\linewidth}{!}{
\setlength\tabcolsep{3.8pt}
\begin{tabular}{c||c|c|c}
\hline \thickhline
\rowcolor{mygray}
\textbf{T-stage} & \textbf{Total} & \textbf{Positive invasion} & \textbf{Negative/absent} \\
\hline\hline
T1 & 151 & 20 (13.2\%) & 131 (86.8\%) \\
T2 & 116 & 25 (21.6\%) & 91 (78.4\%) \\
T3 & 99 & 52 (52.5\%) & 47 (47.5\%) \\
T4 & 73 & 50 (68.5\%) & 23 (31.5\%) \\
\hline \thickhline
\end{tabular}
}
\end{table}

\noindent\textcolor{red}{\textbf{Report Quality and Cross-Institutional Robustness.}}
\textcolor{red}{All radiology reports in our study are routine clinical documents, not research-curated texts. The 439 FUZH training reports vary in detail and phrasing across radiologists while sharing a common reporting template, and the three external test sets (CGMH, EENT, TCGA) follow their own institutional conventions. Among the three modules, AOG operates entirely on CT features and is report-independent; OCA aligns reports to organs through learned cross-attention rather than keyword matching; and REG extracts structural invasion relations from report text that are typically documented even in brief reports when clinically relevant. The stable cross-institutional performance (Table~\ref{tab:compare_sota_vila}) \revB{is consistent with the design accommodating} reporting variation in practice.}

\revB{A controlled report-detail perturbation experiment (Fig.~\ref{fig:robustness}(b)) provides direct support for this: keeping the same images and the frozen model while rewriting each report to be less- or more-detailed changes macro-F1 by at most 1.7 points and accuracy by at most 1.2 points.}

\noindent\textcolor{red}{\textbf{Clinical Deployment Considerations.}}
\textcolor{red}{Several practical challenges should be considered for clinical deployment. First, the framework depends on a pre-trained organ segmentation model, whose performance may vary across imaging protocols and scanner vendors; validating or fine-tuning the segmentation model on site-specific data is advisable before deployment. Second, differences in reporting terminology and detail across institutions may affect the OCA and REG modules, though our cross-institutional evaluation suggests the learned cross-attention mechanism accommodates such variation reasonably well. Third, prospective validation on independent cohorts with different disease distributions is essential before integration into clinical decision-support workflows.}

\section{Conclusion}
In this work, we introduced an anatomy-structured multimodal framework for laryngopharyngeal cancer T-staging that integrates CT volumes and paired radiology reports. By explicitly modeling anatomical topology through the Anatomy-Structured Organ Graph (AOG), performing organ-level cross-modal alignment via the Organ-Anchored Cross-Modal Alignment (OCA) module, and incorporating structured clinical invasion knowledge using the Report-Enhanced Graph-Refinement  (REG) module, our method effectively captures organ-specific invasion patterns and their textual evidence. Extensive experiments on the largest curated laryngopharyngeal cancer T-staging dataset, as well as on public benchmarks, demonstrated that our proposed framework consistently outperforms state-of-the-art unimodal and multimodal baselines in both binary and multi-class staging tasks, achieving notable gains over radiologist assessments. Beyond performance, our framework offers improved interpretability by aligning decision-making with clinically meaningful invasion pathways, thereby enhancing its potential for real-world adoption.
Future work will explore extending the proposed topology-aware design to other head-and-neck and thoracic malignancies, as well as integrating longitudinal imaging and follow-up reports to enable dynamic tumor progression modeling.

\appendices

{\small
\bibliographystyle{IEEEtran}
\bibliography{main}
}

\end{document}